\documentclass[11pt]{article}

\usepackage[a4paper,margin=1in]{geometry}

\usepackage[T1]{fontenc}
\usepackage[utf8]{inputenc}

\usepackage{lmodern}        
\usepackage{microtype}      

\usepackage{booktabs}

\usepackage{color}
\usepackage{latexsym}              
\usepackage{amsmath}               
\usepackage{amssymb}               
\usepackage{amsfonts}              
\usepackage{amsthm}                
\usepackage{multirow}
\usepackage{tikz}                  
\usetikzlibrary{arrows,positioning,shapes}
\usepackage{tcolorbox}

\usepackage{algorithm}
\usepackage{algpseudocode}

\usepackage[round]{natbib}

\definecolor{myblue}{RGB}{25,55,109}
\RequirePackage[%
  pdfstartview=FitH,%
  breaklinks=true,%
  bookmarks=true,%
  colorlinks=true,%
  linkcolor= myblue,
  anchorcolor=myblue,%
  citecolor=myblue,
  filecolor=myblue,%
  menucolor=myblue,%
  urlcolor=myblue%
  ]{hyperref}

\usepackage{cleveref}

\theoremstyle{plain}  
\newtheorem{theorem}{Theorem}[section]

\newcommand{\zbf}{\ensuremath{\mathbf{z}}}

\newcommand{\Dcal}{\ensuremath{\mathcal{D}}}

\newcommand{\Hcal}{\ensuremath{\mathcal{H}}}

\newcommand{\Mcal}{\ensuremath{\mathcal{M}}}
\newcommand{\Ncal}{\ensuremath{\mathcal{N}}}

\newcommand{\Scal}{\ensuremath{\mathcal{S}}}

\newcommand{\Ebb}{\ensuremath{\mathbb{E}}}

\newcommand{\Nbb}{\ensuremath{\mathbb{N}}}

\newcommand{\KL}{\operatorname{KL}}

\newcommand{\LP}{\left(}
\newcommand{\RP}{\right)}

\newcommand{\ie}{{\it i.e.}}

\newcommand{\eg}{{\it e.g.}}

\newcommand{\iid}{{\it i.i.d.}}

\newcommand{\defeq}{:=}

\DeclareMathOperator*{\EE}{\Ebb}

\renewcommand{\H}{\Hcal}
\newcommand{\h}{h}

\newcommand{\N}{\Nbb}

\renewcommand{\P}{\mathrm{P}}
\newcommand{\Q}{\mathrm{Q}}
\renewcommand{\S}{\Scal}
\newcommand{\Risk}{\text{R}}

\newcommand{\z}{\zbf}

\title{Federated Learning with Nonvacuous Generalisation Bounds}

\author{Pierre Jobic \thanks{Université Paris-Saclay CEA, France \texttt{pier.jobic@gmail.com}}
    \and
    Maxime Haddouche \thanks{Inria, CNRS, Ecole Normale Supérieure, PSL Research University, France. \texttt{maxime.haddouche@inria.fr}}
    \and 
    Benjamin Guedj \thanks{Inria, University College London \texttt{b.guedj@ucl.ac.uk}}
      }

\date{}

\begin{document}

\maketitle

\abstract{We introduce a novel strategy to train randomised predictors in federated learning, where each node of the network aims at preserving its privacy by releasing a local predictor but keeping secret its training dataset with respect to the other nodes. 
  We then build a global randomised predictor which inherits the properties of the local private predictors in the sense of a PAC-Bayesian generalisation bound. We consider the synchronous case where all nodes share the same training objective (derived from a generalisation bound), and the heterogenous and homogenous cases where each node may have its own personalised training objective. We show through a series of numerical experiments that our approach achieves a comparable predictive performance to that of the batch approach where all datasets are shared across nodes. Moreover the predictors are supported by numerically nonvacuous generalisation bounds while preserving privacy for each node. We explicitly compute the increment on predictive performance and generalisation bounds for our two federated settings, highlighting the price to pay to preserve privacy.}


\section{Introduction}





In the federated learning (FL) paradigm, a group of \emph{users} (or nodes) is learning in parallel, and typically aims at preserving their personal datasets while sharing a common predictor. While maintaining the privacy of their own data, users mutually share information through a central \emph{server}. There has been a significant surge of interest in federated learning in the past decade~\cite{konevcny2016federated}, with clear applications in healthcare, transportation and retail, where it is typically of the utmost interest to avoid the leak of private information to other organisations or devices, for ethical or business motivations. The existing literature essentially categorises \emph{horizontal} and \emph{vertical} FL, depending on whether users' datasets share many features or individuals. These two streams have generated various contributions such as the design of efficient communication
strategies \cite{konevcny2016fedop,konevcny2016federated,suresh2017distri}, the preservation of privacy through
differentially-private distributed optimisation methods \cite{agarwal2018cpsgd}, the enforcement of fairness (as in many cases, post-training learning models may be biased or
unfair and may discriminate against some protected groups -- \cite{hardt2016equal,mohri2019agnostic}). We refer to \cite{zhang2021survey,mammen2021federated,kairouz2021fed} for recent surveys on FL.

Consider a simple federated learning framework \cite{bonawitz2017practical,mcmahan2023communicationefficient}. In each round, the server first provides the
initial model to each user, then each user updates the initial model with its personal data. Finally, the server aggregates the collected local
models into a single global model, which is used as next round's initialisation if needed. Hereafter we will refer to this learning problem as \emph{FL-SOB} (Federated Learning with Synchronous OBjectives). This is especially relevant when all users share a common learning goal (\eg, hospitals learning from different datasets to identify or predict a specific single pathology). Deep neural networks have been used to develop powerful federated algorithms \cite{mcmahan2017comm}. 
A more complex scenario consists in \emph{personalised FL} (\emph{PFL}, \cite{tan2021personalised}) where users may have their own distinct learning goal but still want to share joint information as these goals share some level of similarity. This corresponds, for instance, to \emph{transfer learning} (see \eg, \cite{zhuang2021survey}) situations where one wants to extract some information of a learning problem (\eg, detecting tigers in images) to perform better on another one, sharing some similarities (\eg, detecting cats). 

\textbf{Towards a unified framework.} The recent PAC-FL framework of \cite{zhang2023probably} proposes a unified framework to formalise FL, intricating the notion of generalisation ability (designed as utility) alongside privacy, and quantifying how much data are protected (\emph{i.e.} impossible to retrieve) while transmitting partial information to the server.  
This framework builds from the work of \cite{zhang2019theo} investigating the tradeoffs between privacy, utility and efficiency. The question of an optimal trade-off is crucial to deploy the FL framework in practice \cite{tsipras2019robustness}.

\textbf{On the place of generalisation in FL.} Using their PAC-FL framework, \cite{zhang2023probably} proposed generalisation bounds involving the dimension of the predictor space. The question of generalisation in FL is central: \cite{mohri2019agnostic,zhang2023federated} established Rademacher-based generalisation bounds, \cite{yagli2020info,barnes2022improved,chor2023more} provided bounds based on mutual information to explain both generalisation ability and privacy leakage per user. \cite{wang2025gene} managed to get generalisation bounds tailored to the FL setting via the conditional mutual information framework, a toolbox also involved in \cite{kavian2025heterogeneity} to highlight the impact of data heterogeneity. Bayesian methods have also been considered in FL-SOB \cite{yurochkin2019bayesian,chen2021fedbe,zhang2022personalised} as well as in PFL \cite{kotelevskii2022fedpop}. Beyond Bayesian methods, \cite{sun2024understanding} involved algorithmic stability to highlight the impact of data heterogeneity, \cite{hu2023gene} involved the VC-dimension of the predictor class to reach generalisation bounds distinguishing participating from unparticipating clients.

\textbf{PAC-Bayes learning in FL.} Beyond Bayesian methods, PAC-Bayes learning (see the seminal works of \cite{ShaweTaylorWilliamson1997,McAllester1998,McAllester2003,maurer2004note} -- we refer to the surveys of \cite{guedj2019primer,alquier2021userfriendly}, and to the recent monograph of \cite{hellstrom-23a}) has recently re-emerged as a powerful framework in batch learning to explain the generalisation ability of neural nets by providing non-vacuous generalisation bounds \cite{DziugaiteRoy2017,letarte19dichotomize,perezortiz2021tight,biggs2020differentiable,biggs2022shallow}. PAC-Bayes combines information-theoretic tools with the Bayesian paradigm of generating a data-dependent \emph{posterior}
distribution over a predictor space from a \emph{prior} distribution (or reference measure), usually data-independent. The flexibility of the PAC-Bayes framework makes it useful to explain generalisation in many learning settings. In particular, theoretical results and practical algorithms have been derived for various learning problems such as reinforcement learning \cite{fard2010pac}, online learning \cite{li2018,HaddoucheGuedj2022}, constrative learning \cite{nozawa2019pacbayesian}, generative models \cite{cherief2021vae}, multi-armed bandits \cite{seldin2011pac,seldin2012pac,sakhi2022pacbayesian}, meta-learning \cite{AmitMeir2018,FaridMajumdar2021,rothfuss2021pacoh,rothfuss2022pac,DingChenLevinboimGoodmanSoricut2021}, majority votes \cite{zantedeschi2021learning,biggs2022majority} to name but a few.

Recently, some works have used the PAC-Bayes framework in FL: \cite{reisizadeh2020robust} and \cite{achituve2021perso} have evaluated the post-training predictor shared by all users through a PAC-Bayes bound. Rather than exploiting existing bounds, new PAC-Bayes results, tailored for personalised FL, recently emerged with the aim to explain the efficiency of learning procedures \cite{scott2023pefll,sefidgaran2023fed}, although the PAC-Bayes bound is not minimised by the algorithm. Finally, recent works showed that the Bayesian procedure ELBO, adapted to FL, is exactly the minimisation of a PAC-Bayes upper bound \cite{kim2023fedhb,vedadi2023fed}. Thus, they show that those methods are well incorporated in a theoretical framework explaining their good generalisation ability. 

A close approach from ours lies in \cite{zhang2024improving}. In their work, they provide novel FL algorithms based on estimating a Gibbs posterior for each client and then aggregating them through the server. The choice of Gibbs posteriors is deeply linked to PAC-Bayes learning as it is the minimiser of Catoni's bound \cite{Catoni2007,alquier2021userfriendly}. Their theoretical backbone is a mutual information bound which can be interpreted as an expected-- yet not computable-- version of classical PAC-Bayes bounds, holding with high probability. Alternatively, we propose here FL algorithms based on high-probability PAC-Bayes bounds and we choose to restrict the optimisation on the set of diagonal Gaussian measures. Those choices allow the practical computation of our bounds. This is a significant difference between our work and theirs as their expected bounds are not computable (as the expectation involve the unknown data distribution). This strategy of deriving learning algorithms from PAC-Bayes bounds also appeared in \cite{zhao2024fed} for non-iid data. They also did not compute theoretical guarantees.

 \subparagraph{Our contributions.} Beyond being a safety check for generalisation, PAC-Bayes theory provides state-of-the art learning algorithms with tight generalisation guarantees in the batch setting. 
 We adapt those algorithms to the FL-SOB and PFL settings. We propose \textsc{GenFL} (standing for Generalisation-driven Federated Learning), 
 an algorithm in which users optimise local PAC-Bayes objectives (bounds from \cite{DziugaiteRoy2017,perezortiz2021tight}). 
 We show a global generalisation bound for all users in FL-SOB, and local ones in PFL.  
 Finally, we show in numerical experiments that our procedure is competitive with the state-of-the-art and we bring nonvacuous generalisation guarantees to practitioners of federated learning. 

 \vspace{1mm}
 \textbf{Outline.} We describe our notation in \Cref{sec:framework} and introduce in \Cref{sec:GenFL} a novel algorithm called \textsc{GenFL}, alongside two instantiations to FL-SOB and PFL. We present numerical experiments to support our methods. in \Cref{sec:expes}. Our algorithms and the code used to generate figures in this paper is available at \url{https://github.com/PierreJobic/GenFL}. The paper closes with a discussion in \Cref{sec:discussion}. 
In \Cref{sec: kl_calculus}, we comment on strategies to compute PAC-Bayesian bounds, \Cref{sec: details_personalised} contains a comprehensive description of our procedure in the PFL setting, and \Cref{sec:additional_expes} provides additional experiments.

\section{Background}
\label{sec:framework}
\textbf{Federated learning.} We consider a predictor set $\Hcal$, a data space $\mathcal{Z}$ and denote the space of distributions over $\mathcal{H}$, $\mathcal{M}(\Hcal)$. We let $\ell \colon \mathcal{H}\times \mathcal{Z} \rightarrow [0,1]$ denote a loss function.
In FL, we consider an ensemble of $K\in\Nbb^*$ users, and for each user $1\leq i \leq K$, we denote by $\S_i =(\z_{i,j})_{j=1\cdots m_i}$ its associated dataset of size $m_i$. We define $\S$, of size $m=\sum_{i=1}^K m_i$, to be the union of all $\S_i$.  We assume that each $\S_i$ is \iid with associated distribution $\mathcal{D}_i$.
Each user $1\leq i \leq K$ aims to jointly learn a predictor $h\in\Hcal$ while keeping private their training dataset $\S_i$. 

\vspace{1mm}
\textbf{Learning theory.} In PAC-Bayes learning, instead of directly crafting a predictor $h\in\Hcal$, we design a data-driven posterior distribution $\Q\in\mathcal{M}(\Hcal)$ with respect to a prior distribution $\P$. 
To assess the generalisation ability of a predictor $h\in\Hcal$, we define for each user $i$ the \emph{risk} to be $\Risk_{\Dcal_i}\defeq \EE_{\z\sim \Dcal_i}[\ell(h,\z)]$ and its empirical counterpart $\hat{\Risk}_{\S_i} \defeq \frac{1}{m_i}\sum_{j=1}^{m_i} \ell(h,\z_{i,j})$. As PAC-Bayes focuses on elements of $\mathcal{M}(\Hcal)$, we also define the expected risk and empirical risks for $Q\in\mathcal{M}(\Hcal)$ as $\Risk_{\Dcal_i}(\Q):= \EE_{\h\sim \Q}[ \Risk_{\Dcal_i}(\h)]$ and $\hat{\Risk}_{\S_i}(\Q):= \EE_{\h\sim \Q}[ \hat{\Risk}_{\S_i}(\h)]$. 

\vspace{1mm}
\textbf{Background on PAC-Bayes learning.}
In a batch setting, we only consider the dataset $\S$ (this can be seen as the case where there is only one user) and we assume that all data are \iid with distribution $\Dcal$. For two probability measures $\P,\Q$ we define the \emph{Kullback-Leibler divergence} to be $\KL(\Q,\P) = \mathbb{E}_{h\sim \P} \left[ \frac{d\Q}{d\P}(h)  \right]$ where $\frac{d\Q}{d\P}$ is the Radon-Nikodym derivative. We also denote by $\mathrm{kl}$ the KL divergence between two Bernoulli distributions.

\subparagraph{Generalisation bounds.} We recall the following bound, due to \cite{McAllester2003,maurer2004note}.

\begin{theorem}[Langford-Seeger-Maurer's bound \cite{seeger2002pac,maurer2004note}]
\label{th:mcall}
For any data-free prior distribution $\P \in \Mcal(\H)$, $\ell\in[0,1]$, $\delta\in [0,1]$, with probability at least $1-\delta$, for any posterior distribution $\Q\in\Mcal(\Hcal)$,
\begin{align}
\label{eq:mcall_tight}
\mathrm{kl}\left(\Risk_{\Dcal}(\Q), \Risk_\S(\Q)  \right) \leq \frac{\KL(\Q\|\P) + \ln{\frac{2\sqrt{m}}{\delta}}}{m},
\end{align}
which leads to the following upper bound on the risk
\begin{align}
\label{eq:mcall_invert}
\Risk_{\Dcal}(\Q) \le \mathrm{kl}^{-1} \left(\hat{\Risk}_{\S}(\Q) \left\| \frac{\KL(\Q\|\P) + \ln{\frac{2\sqrt{m}}{\delta}}}{m}\right)\right.,
\end{align}
where  $\mathrm{kl}^{-1}(x,b)= \sup\left\{ y\in [x,1] \mid kl(x,y) \leq b \right\}$. 
\end{theorem}

Note that by the definition of $kl^{-1}$, \eqref{eq:mcall_invert} is the tightest upper bound on $\Risk_\Dcal(\Q)$ that we can obtain starting from \eqref{eq:mcall_tight}. While $\mathrm{kl}^{-1}$ has no closed form, it is possible to approximate it efficiently via root-finding techniques (see, \eg, \cite[Appendix A]{DziugaiteRoy2017}). However, this function is hard to evaluate, and even harder to optimise. We need to rely on looser relaxations of \eqref{eq:mcall_tight} to design tractable optimisation procedures. 

\vspace{1mm}
\textbf{McAllester's bound.} The most classical relaxation of \eqref{eq:mcall_tight} relies on Pinsker's inequality $\mathrm{kl}(q\|p)\geq \frac{(p-q)^2}{2}$ and leads to the following high-probability bound, valid under the same assumptions as \Cref{th:mcall}:
\begin{align}
\label{eq:mcall_basic}
\Risk_{\Dcal}(\Q) \le \hat{\Risk}_{\S}(\Q) + \sqrt{ \frac{\KL(\Q\|\P) + \ln{\frac{2\sqrt{m}}{\delta}}}{2m}}.
\end{align}

While \eqref{eq:mcall_basic} is well known and already appears in \cite{McAllester2003}, novel relaxations, exploiting refined Pinsker's inequality (see, \eg, \cite[Lemma 8.4]{boucheron2013conc}) have been exploited to obtain PAC-Bayes Bernstein bounds \cite{tolstikhin2013pac,mhammedi19pac}. Building on this inequality, \cite{rivasplata2019back,perezortiz2021tight} proposed a \emph{PAC-Bayes quadratic bound} recalled below, valid under the assumptions of \Cref{th:mcall}
\begin{align}
\label{eq:mcall_quad}
\Risk_\Dcal(\Q) & \leq\left(\sqrt{\hat{\Risk}_\S(\Q)+\frac{\mathrm{KL}\left(\Q \| \P\right)+\log \left(\frac{2 \sqrt{m}}{\delta}\right)}{2 m}} +\sqrt{\frac{\KL\left(\Q \| \P\right)+\log \left(\frac{2 \sqrt{m}}{\delta}\right)}{2 m}}\right)^2.
\end{align}

Note that \eqref{eq:mcall_basic} and \eqref{eq:mcall_quad} are easier to optimise than \eqref{eq:mcall_invert}, making them more relevant for practical learning algorithms.

\subparagraph{Generalisation-driven learning algorithms.} Most of the PAC-Bayesian bounds in the literature are fully empirical. This paves the way to use the bound as a training objectives and leads to generalisation-driven learning algorithms. 
A classical PAC-Bayesian algorithm is derived from Catoni's bound (see, \eg, \cite{Catoni2007}, \cite[Theorem 4.1]{AlquierRidgwayChopin2016}):
\begin{align}
\label{eq:catoni_alg}
\operatorname*{argmin}_{\Q \in \Mcal_(\Hcal)} \Risk_\S(\Q) +\frac{\KL(\Q,\P)}{\lambda}.
\end{align}
In \eqref{eq:catoni_alg} an \emph{inverse temperature} $\lambda>0$ appears and acts as a learning rate in gradient descent. Similarly, it is possible to derive batch learning algorithms from \eqref{eq:mcall_basic} \eqref{eq:mcall_quad} \cite{DziugaiteRoy2017,perezortiz2021tight}. Then, we have access to a theoretical upper bound which requires to approximate expectations over $\Q$. We next discuss how to mitigate this.

\vspace{1mm}
\textbf{Computing generalisation guarantees.}
In practice, we compute the Langford-Seeger-Maurer's bound \eqref{eq:mcall_tight}. First, note that the KL divergence is easy to compute in the Gaussian case as it has a closed form (see, \eg, \cite[Section 9]{duchi2007derivations}). Then, it remains to estimate the expected empirical risk over $\Q$, which is costly in practice as it involves Monte Carlo approximations. To alleviate this issue, we leverage the trick from \cite[Section 3.3]{DziugaiteRoy2017}, which exploit a high probability upper bound of $\hat{\Risk}_\S(\Q)$ with confidence level $\delta'$
\begin{align*}
\hat{\Risk}_{\S}(Q) & \le \mathrm{kl}^{-1} \left(\hat{\Risk}_{\S}(\hat{Q}_n) \left\| \frac{1}{n} \ln\LP\frac{2}{\delta^{'}}\RP\right)  \right., 
\end{align*}
where $n$ is the number of Monte-Carlo samples, $\hat{Q}_n = \frac{1}{n}\sum_{i=1}^n \delta_{W_i}$ and $\hat{\Risk}_{\S}({\hat{Q}_n}) = \frac{1}{n} \sum_{i=1}^n \hat{\Risk}_{\S}(W_i)$, where $W_i\sim Q$ are iid and $\delta$ denotes the Dirac distribution. Estimating $\hat{R}_S(\hat{Q}_n)$, requires $n\times m$ computation of the loss.

Then incorporating this upper bound in \eqref{eq:mcall_tight} gives the final bound we use, with probability at least $1-\delta-\delta'$:

\begin{align}
     \label{eq:mcall_final}
     \Risk_{\Dcal}(\Q) \le \mathrm{kl}^{-1} \left(\hat{\Risk}_{\S}^{n}(\Q) \left\| \frac{\KL(\Q\|\P) + \ln{\frac{2\sqrt{m}}{\delta}}}{m}\right)\right.,
\end{align}

where $\hat{\Risk}_{\S}^{n}(\Q)=\mathrm{kl}^{-1} \left(\hat{\Risk}_{\S}(\hat{Q}_n) \left\| \frac{1}{n} \ln\LP\frac{2}{\delta^{'}}\RP\right)  \right.$.

To compute $\mathrm{kl}^{-1}(p,c)$, we leverage \cite[Appendix A]{DziugaiteRoy2017} described in \Cref{sec: kl_calculus}.

\section{Generalisation-driven Federated Learning}
\label{sec:GenFL}

In this section we introduce our algorithm \textsc{GenFL}.

\vspace{1mm}
\textbf{From batch to federated PAC-Bayes algorithms.} 
When training stochastic neural nets (SNNs) with PAC-Bayes objectives, it is common to assume that each weight follows a Gaussian distribution. For conciseness, we identify a SNN to the Gaussian distribution of all its weights $\mathcal{N}(\mu,Diag(\sigma_i))$. 
The works of \cite{DziugaiteRoy2017,rivasplata2019back,biggs2020differentiable,perezortiz2021tight,perezortiz2021progress,biggs2022shallow} proposed successful PAC-Bayesian training algorithms for SNNs which ensure generalisation guarantees. All these methods operate in a batch setting, \ie, the optimiser has access to all data simultaneously. Thus, building on the work of \cite{rivasplata2019back,perezortiz2021tight}, we propose \cref{alg:GenFL}, a new learning algorithm called \textsc{GenFL} (Generalisation-driven Federated Learning), casting PAC-Bayes into FL. We stress that \textsc{GenFL} benefits from nonvacuous generalisation guarantees (\Cref{sec:expes}).

\begin{algorithm}[htbp]
\caption{\textsc{GenFL}. users are indexed by $k$; $B$ is the local minibatch size, $E$ is the number of local epochs, $\eta$ is the learning rate, $f$ the PAC-Bayes objective. The prior is $\Ncal(\mu_{\text{prior}}\sigma_{\text{prior}})$ with parameter $\delta$.}\label{alg:GenFL}
\begin{algorithmic}[1]
\State \textbf{Server executes:}
\State $m \leftarrow \sum_{k=1}^{K} m_k$ \Comment{Total dataset size}
\State $w_1 \leftarrow (\mu_{\text{prior}}, \sigma_{\text{prior}})$
\For{each round $t$}
\State $S_t \leftarrow \text{random set of } \max(C \cdot K, 1) \text{ users}$ \Comment{$C\in(0,1)$ is the proportion of clients per round}
\For{each user $k \in S_t$ \textbf{in parallel}}
\State $w_{t+1}^k \leftarrow \textrm{userUpdate}(k, w_t, m)$
\EndFor
\State $w_{t+1} \leftarrow \sum_{k=1}^{K} \frac{m_k}{m} w_{t+1}^k$
\EndFor
\State
\State \textbf{userUpdate(k, $w$, m):}
\State $w^k \leftarrow w \quad \triangleright~\text{initialise local }(\mu_k,\sigma_k)\text{ from global }(\mu_t,\sigma_t)$
\State $\mathcal{B} \leftarrow (\text{split } \S_k \text{ into batches of size B})$
\For{each local epoch $e= 1, 2, \cdots, E$}
\For{each local minibatch $b \in B$}
\State $w^k_s \leftarrow \mu^k + \sigma^k \odot \mathcal{N}(0,1)$ \Comment{Reparam. trick}
\State $w^k \leftarrow w^k - \eta \nabla_w f_{m, \delta, \mu_{\text{prior}}, \sigma_{\text{prior}}}(w^k_s; b)$ \Comment{$\sigma^k \leftarrow \mathrm{softplus}(\sigma^k) = \ln(1+e^{\sigma^k})$ \text{ to ensure } $\sigma^k>0$.}
\EndFor
\EndFor
\State \textbf{return} $w^k$
\Ensure \textbf{Output: } $Q=\mathcal N(\mu_T,\sigma_T)$
\end{algorithmic}
\end{algorithm}

\textbf{A generalisation-driven FL algorithm.} \textsc{GenFL} combines a federated learning 
protocol (\ie, FedSGD, FedAvg, \cite{mcmahan2017comm}) with a PAC-Bayes objective $f$. Its starts from the vector $w_1= (\mu_{prior},\sigma_{prior})$ corresponding to the initial distribution $\P= \Ncal(\mu_{prior},\sigma_{prior})$ and outputs after $T$ rounds $w_T=(\mu_T,\sigma_T)$ corresponding to the posterior $Q=\Ncal(\mu_T,\sigma_T)$. Hence the learning procedure is divided in rounds, where subsets of users are sampled. 
Sampled users are requested to perform local updates following a PAC-Bayes training procedure designed to ensure a good generalisation of the posterior distribution. Because users train SNNs, they sample weights from the posterior. In order to learn the variance parameter of the posterior (Gaussian), we use the well-known reparameterisation trick \cite{kingma2013autoencoding}: instead of directly sampling from the distribution, we sample from a standard Gaussian distribution and then apply a transformation to obtain the sampled weights: $W = \mu + \sigma V$ where $V \sim \mathcal{N}(0, \mathrm{Id})$, this allows to compute with respect to $\sigma$.
When the round ends, the global model is computed from the local updates, thanks to the aggregation function from the FL protocol (weighted mean, median). 

Next we show that with a PAC-Bayes objective $f$, we adapt \textsc{GenFL} to FL-SOB, where $\S$ is fully \iid, \ie, all user datasets have the same distribution, and to PFL where each dataset $\S_i$ is \iid but any two dataset can have distinct distributions.   

\subsection{\textsc{GenFL} for FL-SOB} 
\label{sec:basic_fl}
\textbf{PAC-Bayesian objectives.} We assume that for any $i$, $\Dcal_i= \Dcal$. Thus, $\S$ is a \iid dataset of $m$ points. We then consider the true and empirical risks on all $\S$ $\Risk_\Dcal, \Risk_{\S}$.
In a batch setting, it would be natural to optimise the bounds \eqref{eq:mcall_basic}, \eqref{eq:mcall_quad}. However, the user $i$ only has access to its personal dataset $\S_i$ (of size $m_i$) to optimise its model \emph{while knowing other datasets are involved}. We then derive accordingly from \eqref{eq:mcall_basic}, \eqref{eq:mcall_quad} two PAC-Bayesian learning algorithms, valid for any user, namely $f_1,f_2$. Note that $f_1$ is adapted from the $f_{classic}$ objective and $f_2$ is adapted from $f_{quad}$ of \cite{perezortiz2021tight}:
\begin{align}
\label{eq:obj_classic_ordered}
f_{1}(\S_i) & =  \hat{\Risk}_{\S_i}(Q) + \sqrt{ \frac{\KL(Q\|P) + \ln{\frac{2\sqrt{m}}{\delta}}}{2m}}. \\
\label{eq:obj_quad_ordered}
f_{2}(\S_i) & = \left(\sqrt{\hat{\Risk}_{\S_i}(Q)+\frac{\mathrm{KL}\left(Q \| P\right)+\log \left(\frac{2 \sqrt{m}}{\delta}\right)}{2 m}} +\sqrt{\frac{\KL\left(Q \| P\right)+\log \left(\frac{2 \sqrt{m}}{\delta}\right)}{2 m}}\right)^2.
\end{align}
Note that \eqref{eq:obj_classic_ordered}, \eqref{eq:obj_quad_ordered} can be seen as proxys of \eqref{eq:mcall_basic}, \eqref{eq:mcall_quad}. Indeed, every user has access to the total number of data points $m$ (as long as it is transmitted to the server), so the regularisation term (containing the KL divergence) is fully available, contrary to the empirical risk $\hat{\Risk}_{\S}$ which is then replaced by $\hat{\Risk}_{\S_i}$. Note that in this case, the  KL divergence is divided by $m$ instead of $m_i$ (which would be natural if we were optimising \eqref{eq:mcall_basic} \eqref{eq:mcall_quad} for $\S_i$ instead of $\S$). This suggests that each user has to give more weight, during the optimisation phase, to its data than to the regularisation. The reason behind this is that the server, by aggregating predictors, performs a regularisation step on a global level, hence the need to prioritise data on a local one. 

\vspace{1mm}
\textbf{A global generalisation guarantee.} A major interest of the \iid assumption on $\S$ is that, as long as users all exploit the same posterior distribution $Q$, and they transmit their empirical scores $\hat{\Risk}_{\S_i}(Q)$, every user is able to compute the global generalisation guarantee of \eqref{eq:mcall_invert}. This allows to maintain the bound of the batch setting (involving the total number of data points $m$), despite being in FL. This is empirically shown in \Cref{sec:expes}. We present in \Cref{alg:FedBound} \textsc{Fedbound}, the algorithm we use to compute the global bound \eqref{eq:mcall_final}, valid for all users simultaneously.

\begin{algorithm}[htbp]
\caption{\textsc{FedBound}. The K users are indexed by k; $f$ PAC-Bayes objective, prior $\Ncal(\mu_{\text{prior}}$, $\sigma_{\text{prior}})$; posterior $\Ncal(\mu_{T}$, $\sigma_{T})$, $\delta$, $\delta^{'}$ parameters, $n$ number of Monte Carlo sampling}
\label{alg:FedBound}
\begin{algorithmic}[1]
\State \textbf{Server executes:}
\State $m \leftarrow \sum_{k=1}^{K} m_k$ \Comment{Total dataset size}
\State $\P = \mathcal{N}(\mu_{\text{prior}}, \sigma_{\text{prior}})$ \Comment{Prior (learned or random)}
\State $\Q = \mathcal{N}(\mu_{T}, \sigma_{T})$ \Comment{Posterior (learned by \Cref{alg:GenFL})}
\For{each user $k \in K$ \textbf{in parallel}}
\State $error^{k} \leftarrow \textrm{userMCSampling}(k, w_t, m)$
\EndFor
\State $error \leftarrow \sum_{k=1}^{K} \frac{m_k}{m} error_k$
\State $KL\_inv \leftarrow \mathrm{kl}^{-1} \left(error \mid \frac{1}{n} \ln(\frac{2}{\delta^{'}})\right)$
\State $\texttt{Up-bound} \leftarrow \mathrm{kl}^{-1} \left(KL\_inv \mid \frac{\KL(Q\|P) + \ln{\frac{2\sqrt{m}}{\delta}}}{m}\right)$
\State
\State \textbf{userMCSampling(k, w, m):}
\For{each MC sampling $i= 1, 2, \cdots, n$}
\State $W^k_i \sim \Q$ \Comment{Sample weights from the posterior}
\State $error^{k}_i \leftarrow \hat{\Risk}_{\S_k}(W^k_i)$ \Comment{local empirical risk}
\EndFor
\State $error^{k} \leftarrow = \frac{1}{n} \sum_{i=1}^{n} error^{k}_i$
\State \textbf{return} $error^{k}$
\Ensure \texttt{Up-bound} holding with probability $1- \delta - \delta^{'}$
\end{algorithmic}
\end{algorithm}

In \Cref{alg:FedBound} we set the prior $P=\mathcal N(\mu_{\text{prior}},\sigma_{\text{prior}})$ and the posterior $Q=\mathcal N(\mu_T,\sigma_T)$. 
Here, $Q$ is the posterior \emph{learned by \Cref{alg:GenFL}} after $T$ federated rounds, while $P$ is either random (data-free) or data-dependent (learned) as described in Sec.~4.1.1 (see also \Cref{alg:p-FL_alg}).

\subsection{\textsc{GenFL} for personalised federated learning} \label{sec:personalised_fl}

\textbf{A general training for the prior distribution.} In PFL, the learning objective of each user may differ, while sharing some similarities that can be learned and transferred from one user to another. This framework requires adjustments of our learning objectives. Indeed, contrary to \Cref{sec:basic_fl}, there is no clear global generalisation guarantee, so each user has then to optimise its own personal learning objective from a commonly shared prior. Using either a random prior or a learnt one on a fraction of users data, we run \textsc{GenFL} similarly to \Cref{sec:basic_fl} with our PAC-Bayesian objective of interest $f$. The output distribution of \textsc{GenFL} is then considered as a common prior for all users which then needs to be personalised. 

\vspace{1mm}
\textbf{A personalisation step.} Once a common prior distribution has been obtained from the federated training, it is necessary for each user to personalise it to its own problem. To do so, we apply PAC-Bayesian objectives similar to those in \Cref{sec:basic_fl}, namely $f_{1}$ \eqref{eq:obj_classic_ordered} and $f_{2}$ \eqref{eq:obj_quad_ordered}, where the batch size is modified from $m$ to $m_i$ for the $i$-th user. This reflect that each user now optimises its local goal instead of the global one. Each user ends up with its own personalised posterior distribution.
The way personalised bounds are implemented is similar to \Cref{alg:FedBound}, but without the aggregation step. We refer to \Cref{sec: details_personalised} for additional details.

\section{Experiments}
\label{sec:expes}

In this section we provide practical instantiations of \textsc{GenFL}. We first consider in \Cref{sec: expe-MNIST} the case of classification on MNIST in a federated setting using basic neural architectures. We then extend our experimental framework in \Cref{sec:cifar10} to the more challenging case of classification on CIFAR-10 with more sophisticated neural networks.
\subsection{Classification on MNIST}
\label{sec: expe-MNIST}
We first detail our framework and then details our experimental findings.
\subsubsection{Experimental framework}
\label{sec: fram-MNIST}
Our experimental framework is inspired from \cite{perezortiz2021tight} combined with the FedAvg protocol \cite{mcmahan2017comm}.
We use the following libraries: Pytorch \cite{pytorch} for deep learning, Flower \cite{beutel2020flower} for federated learning, Slurm \cite{SLURM} for cluster experiments, and Hydra \cite{Yadan2019Hydra} for overall experiment management. The cluster nodes we use have 48 SKYLAKE 3GHz CPUS. We do not use GPUs.

\vspace{1mm}
\textbf{Prior distribution over weights.}
We propose two types of priors: data-free (random) prior chosen randomly around $\mathcal{N}(0,\mathrm{Id})$ (as in \cite{DziugaiteRoy2017}) and a data-dependent (learnt) prior. The latter is powerful to attenuate the KL divergence term, leading to sharper generalisation bounds and better accuracy.
Such a data-driven prior implies to use a fraction of the dataset from the training data to optimise the prior. The bound computation is then realised with a reduced dataset size (divided by 2 in practice).
However, the prior has gained efficiency (lower empirical risk) and the PAC-Bayes optimisation starts from a relevant point. 

We use Gaussian distribution for both prior and posterior over the weights of a neural network.
When data-free, the prior is $\P = \bigotimes_{l \in \textrm{layers}}\mathcal{N}(\text{truncated}(\mu^l_{\text{rand}}), \textrm{Diag}(\sigma_{\text{prior}}))$  with $\mu^l_{\text{rand}} \sim \mathcal{N}(0, \frac{1}{\sqrt{n^l_{in}}})$, and $\sigma_{\text{prior}} \in \mathbb{R}^{+*}$. The truncature is done at $\pm  \frac{2}{\sqrt{n^l_{in}}}$ where $n^l_{in}$ is the dimension of the inputs of the layer $l$. 
In the case of data-dependent prior, we have $\P = \bigotimes_{l \in \textrm{layers}}\mathcal{N}(\mu^l_{\text{learnt}}, \textrm{Diag}(\sigma_{\text{prior}}))$, where $\mu_{\text{learnt}}$. It is obtained via ERM on the prior set on half of the training set, the other half being used for bound computation and posterior optimisation.

While limiting (and other works may consider richer models such as \emph{Gaussian Mixtures Models (GMMs)} \cite{campos2025federated,pettersson2025federated}), choosing Gaussian distributions with diagonal covariance matrices ensures a tractable KL divergence, thus tractable learning objectives, a classic choice in PAC-Bayes learning \cite{DziugaiteRoy2017,perezortiz2021tight,perezortiz2021progress}. Note also that the Gaussian assumption can be seen as a reasonable proxy for deterministic predictors which are Gaussian distributions with covariance matrix equal to $0\mathrm{Id}$. Alternatively, some works consider directly the Gibs posterior, which is the exact minimiser of Alquier's bound, at the cost of costly computations \cite{Catoni2007,rothfuss2021pacoh,rothfuss2022pac}.

\vspace{1mm}
\textbf{Dataset partition.}
To build a \iid FL setup, we consider the case where each user has exactly the same number of samples per class. 
We partition MNIST as follows: we fix the number of users to 100. Then, each user receives a dataset size of 540, each class having 54 images. 
In the case of the learnt prior, we split the training set of each user in half, the first one being used to train the prior, the second exploited by our learning algorithms. 

In the case of non-\iid FL setup, we follow \cite{mcmahan2017comm}. First we sort MNIST by label, then we partition the dataset into chunks of 300 contiguous samples each (thus containing at most 2 labels, because it is sorted). 
Again we split each user dataset in several parts. When the prior is random, we save 10\% of the dataset to create a validation set. Remaining data is exploited for optimisation. When considering learnt priors, we save again 10\% of the dataset as a validation set, we exploit 40\% of the dataset to train the prior (respecting the proportion of each class), the remaining 50\% being used by our learning algorithms.

\vspace{1mm}
\textbf{Bound parameters.}
We used $\delta=0.05$, $\delta^{'}=0.01$, alongside $n=150 000$ Monte Carlo samples.

\vspace{1mm}
\textbf{Optimisation hyperparameters.}
The prior distribution scale $\sigma_{\text{prior}}$ is set to $2,5\times 10^{-2}$, the learning rate is $5\times10^{-3}$ for 100 users. In order to compare with the batch learning setting, we compute our algorithms with 1 user. In this case, we use a learning rate of $5\times10^{-4}$ to reach better performances.
The momentum is 0.95 for posterior optimisation and 0.99 in prior optimisation.
During prior optimisation, we used a dropout rate of $0.2$ to avoid overfitting. 
As theoretical results of \Cref{sec:framework} require a loss function in $[0, 1]$, we use the bounded cross-entropy as in \cite{perezortiz2021tight}, \ie, $\ell(x, y) = \frac{1}{\ln(p_{\text{min}})}\cdot\ln(\tilde{\sigma}(x)_{y})$ with $\tilde{\sigma}(x)_{y} = \max(\text{sigmoid}(x)_y, p_{\text{min}})$. We took $p_{\text{min}}=10^{-4}$.

\vspace{2mm}
\textbf{KL penalty.} We denote as 'KL penalty' a numerical factor $\alpha\in (0.1)$ such that we optimise $\alpha \KL$ instead of $\KL$ in our learning objectives, \emph{e.g.} \Cref{eq:obj_classic_ordered,eq:obj_quad_ordered}.
For stability reasons, we penalise the KL term during posterior optimisation (similarly to \cite{perezortiz2021tight} or Equation 9 of \cite{zhao2024fed}), thus we give more impact to the empirical risk during optimisation. Such a penalty helps performance and stability during training when random priors are involved. In this case, we use a penalty of $0.1$.

\vspace{1mm}
\textbf{Federated Learning hyperparameters.}
Starting from a random prior, we perform our algorithm during 200 rounds to make the SNNs converge.
When learnt priors are considered, they are trained with a run of 100 rounds with 5 local epochs (convergence around 50 epochs). We then perform 10 additional rounds with 5 local epochs to train the posterior. 
In both cases, we select 10\% of the users each round to participate in the training. 
As the dataset size of each user is small, we use a batch size of 25 (compared to 250 in the work of \cite{perezortiz2021tight}).

\vspace{1mm}
\textbf{Neural Architecture.}
We consider a stochastic 2 hidden layer MLP with 600 units each, resulting in 1,198,210 number of parameters for the prior (with fixed covariance matrix) and doubled for the SNN (as we consider diagonal covariance matrices).

\vspace{1mm}
\textbf{Positive variance prior.}
To have a constrained positive standard deviation $\sigma$ when sampling weights, we use the following transformation: $\sigma = \ln(1 + \exp(\rho))$. 
It makes $\sigma$ always positive, and $\rho$ can be any real number that is optimised during training procedure.

\subsubsection{Results}
Note that in a classification problem, the generalisation error translates a positive influence of the learning phase as long as it is smaller than $1$ (which is what we refer to with the term nonvacuous). Indeed, a bound below this threshold shows that the posterior will not fail at each try. 
However, we focus on posteriors with generalisation bounds or test error smaller than $50\%$. The reason is that, for a binary classification task, this threshold is the generalisation error of a randomised predictor with associated distribution $\mathrm{Bernoulli}(0.5)$. Thus, having results below this threshold provably show we generalise better than a naive strategy.

\vspace{1mm}
\textbf{Federated Learning with Synchronous OBjectives(FL-SOB) setting.}
\begin{table}[bt]
     \caption{ Results for the FL-SOB scenario.
     $\ell^{0-1}$ corresponds to the 0-1 loss. The test error column is made on the test set of MNIST. The
     Bound column corresponds to the generalisation bounds, computed with \cref{alg:FedBound}. The $KL/m$ column corresponds to the KL divergence term in the bound divided by $m=60000$ in data-free prior or $m=30000$ data-dependent prior. }
     \label{tab:PerezOrtiz2020Ordered}
     \vspace{2mm}
     \centering
     \begin{tabular}{|cc|c|c|c|}
     \hline
          \multicolumn{2}{|c|}{Setup}                       & \multicolumn{1}{c|}{Bound}             & \multicolumn{1}{c|}{Test Err.}            & \multicolumn{1}{c|}{KL div}     \\ \hline 
          \multicolumn{1}{|c|}{Prior}       & Obj.          & \multicolumn{1}{c|}{$\ell^{0-1}$}              & \multicolumn{1}{c|}{$\ell^{0-1}$} & \multicolumn{1}{c|}{$KL/m$}   \\ \hline
          \multicolumn{1}{|c|}{\cite{perezortiz2021tight}: Random}     & $f_{1}$ & \multicolumn{1}{c|}{0.330}     & \multicolumn{1}{c|}{0,141}      & \multicolumn{1}{c|}{0,081} 
              \\ 
              \multicolumn{1}{|c|}{(1 client)}       & $f_{2}$    & \multicolumn{1}{c|}{0.316}     & \multicolumn{1}{c|}{0,092}      & \multicolumn{1}{c|}{0,138}  \\ \hline
          \multicolumn{1}{|c|}{\cite{perezortiz2021tight}: Learnt}     & $f_{1}$ & \multicolumn{1}{c|}{0.028}     & \multicolumn{1}{c|}{0,023}      & \multicolumn{1}{c|}{$<$0,001}  \\ 
          \multicolumn{1}{|c|}{(1 client)}      & $f_{2}$    & \multicolumn{1}{c|}{0.028}     & \multicolumn{1}{c|}{0,020}      & \multicolumn{1}{c|}{0,001}  \\ \hline


          
          \multicolumn{5}{|c|}{100 users - GenFL - KL Penalty=0.1} \\ \hline
          \multicolumn{1}{|c|}{Random}        & $f_{1}$ & \multicolumn{1}{c|}{0.333}     & \multicolumn{1}{c|}{0,123}      & \multicolumn{1}{c|}{0,107}   \\ 
          \multicolumn{1}{|c|}{(us)}       & $f_{2}$    & \multicolumn{1}{c|}{0.342}     & \multicolumn{1}{c|}{0,090}      & \multicolumn{1}{c|}{0,163}    \\ \hline
          \multicolumn{1}{|c|}{Learnt}        & $f_{1}$ & \multicolumn{1}{c|}{0,061}     & \multicolumn{1}{c|}{0,030}      & \multicolumn{1}{c|}{$<$0,001}    \\ 
          \multicolumn{1}{|c|}{(us)}      & $f_{2}$    & \multicolumn{1}{c|}{0,088}     & \multicolumn{1}{c|}{0,029}      & \multicolumn{1}{c|}{0,002}    \\ \hline
          
          \multicolumn{5}{|c|}{100 users - GenFL - No KL Penalty} \\ \hline
          \multicolumn{1}{|c|}{Random}        & $f_{1}$ & \multicolumn{1}{c|}{0,415}     & \multicolumn{1}{c|}{0,256}      & \multicolumn{1}{c|}{0,039}\\ 
          \multicolumn{1}{|c|}{(us)}       & $f_{2}$    & \multicolumn{1}{c|}{0,408}     & \multicolumn{1}{c|}{0,251}      & \multicolumn{1}{c|}{0,041} \\ \hline
          \multicolumn{1}{|c|}{Learnt}        & $f_{1}$ & \multicolumn{1}{c|}{0.039}     & \multicolumn{1}{c|}{0,030}      & \multicolumn{1}{c|}{$<$0,001}\\ 
          \multicolumn{1}{|c|}{(us)}      & $f_{2}$    & \multicolumn{1}{c|}{0.040}     & \multicolumn{1}{c|}{0,030}      & \multicolumn{1}{c|}{$<$0,001}  \\ \hline

          \end{tabular}

     \end{table}

\Cref{tab:PerezOrtiz2020Ordered} gathers our results for \textsc{GenFL} applied with $f_{1},f_{2}$ alongside \textsc{FedBound}. Recall that ``KL penalty $=0.1$''means that the KL term in the \emph{training} objective is multiplied by $0.1$. 
Conversely, ``No KL penalty'' uses the full KL with factor $1.0$. 
This weighting affects optimisation but not the KL value later reported inside the PAC-Bayes bound. Also recall that the true generalization error $R_D(Q)$ is a distributional expectation and is not directly observable, we then plotted the empirical generalization error (``test error''), which estimates $R_D(Q)$ with high precision thanks to Hoeffding's inequality. 
Independently, \Cref{alg:FedBound} provides a high-probability \emph{upper bound} on $R_D(Q)$ via the PAC-Bayes bound. We compared our results with 100 clients with the output of our algorithms for 1 client, corresponding to the batch learning case.
Our FL algorithms benefit from nonvacuous theoretical guarantees and test errors. In the case of data-dependent priors, test errors of \textsc{GenFL} nearly reach for both $f_{1}$ and $f_{2}$, the precision of their batch counterpart (3\% in FL and 2\% in batch). In the case of random prior, the KL penalty has a strong positive influence on the test errors. 
The generalisation bounds of our algorithms are uniformly deteriorated compared to the batch setting, \emph{while being of the same magnitude}. This is important to notice as this is the price to pay to adapt batch bounds to a federated setting. Indeed, as each user only optimised a proxy of the common generalisation bound, it is legitimate to retrieve in our results a short discrepancy comparing to the batch case.

While $f_\text{2}$ is consistently achieving better generalisation upper bounds in \cite{perezortiz2021tight}, it is outperformed by $f_{1}$ in the FL setting. However, notice that $f_{2}$ provides uniformly better test errors than $f_{1}$, similarly to \cite{perezortiz2021tight}. Note that better results are achieved if one considers the KL penalty trick with data-free prior and no KL penalty trick with data-dependent prior. 
We interpret this fact as follows: given that the data-dependent prior is already performing well on training data, allowing the posterior optimisation to be unconstrained is not an issue as we found an area close from a local minimiser. However, as the random prior is not necessarily efficient, we need to move far from it to reach good empirical performances. However, moving freely from the prior distribution leads to a large KL divergence, hence the need to constrain the posterior optimisation to obtain both better bounds and test errors. Numerically, for 100 users with a random prior, adding the KL penalty ($\times 0.1$) improves test error: 
$f_1:~0.256 \to 0.123$ and $f_2:~0.251 \to 0.090$ (Table~1). 
We attribute this to the stronger empirical-risk emphasis during training, which helps optimisation when the prior is data-free.
A take-home message is that adapting PAC-Bayes algorithms to FL is effective: it gives nonvacuous results close to the batch setting.

\vspace{1mm}
\textbf{Personalised FL setting.}

\begin{figure}
    \centering
    \resizebox{0.7\textwidth}{!}{%
        \begin{minipage}{\textwidth}
            \centering
            \includegraphics[width=0.45\textwidth]{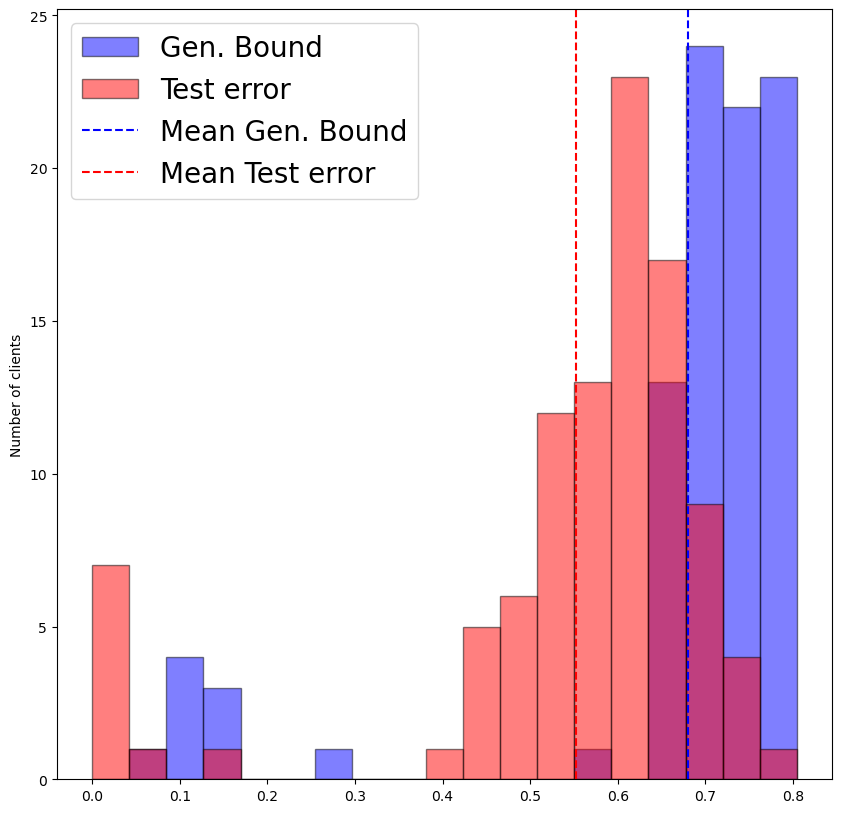}
            \includegraphics[width=0.45\textwidth]{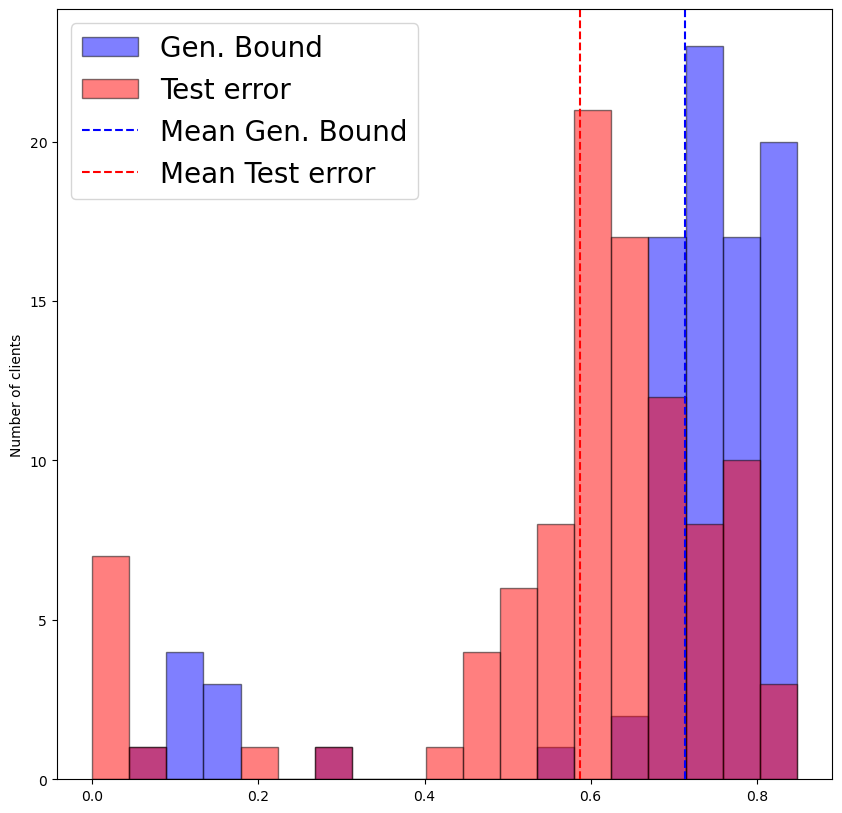}

            \includegraphics[width=0.45\textwidth]{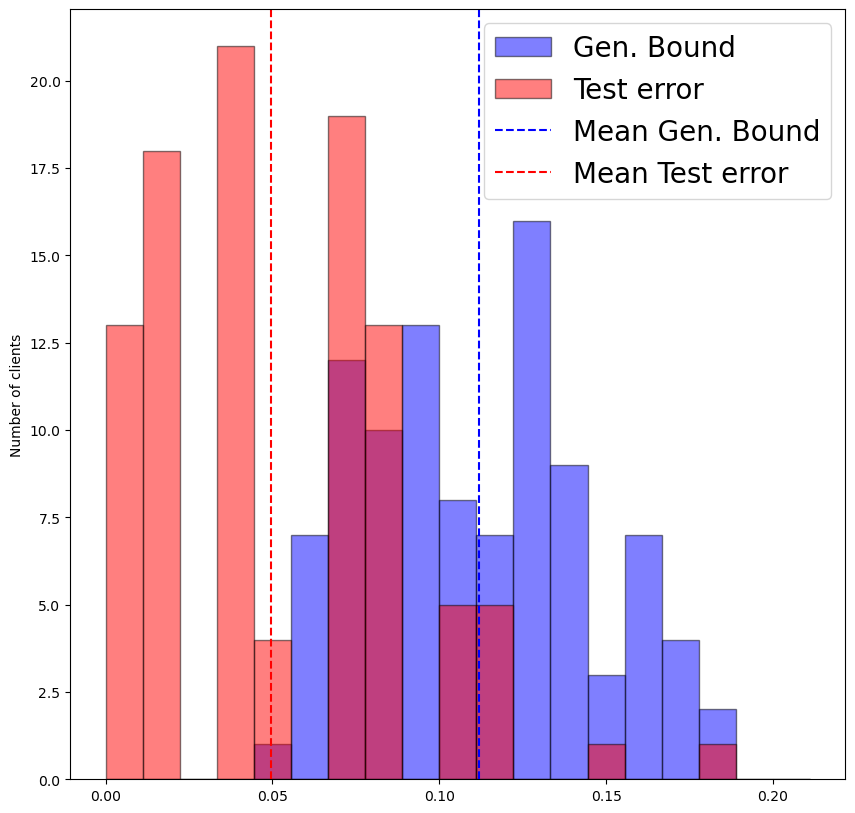}
            \includegraphics[width=0.45\textwidth]{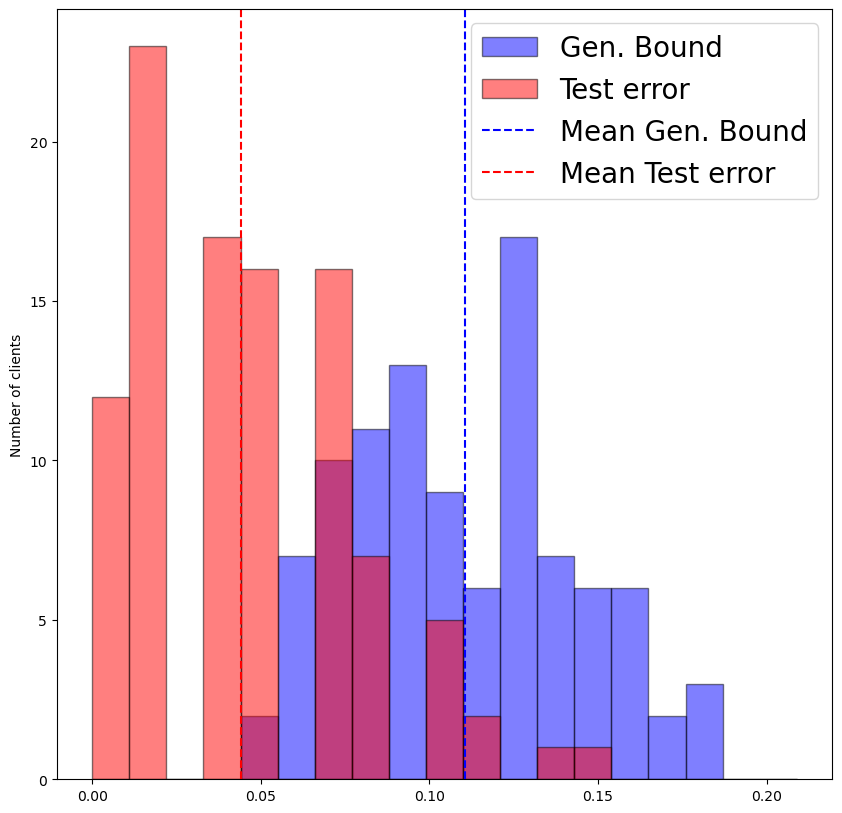}
        \end{minipage}
    }

    \caption{Histograms gathering test errors (red) and bound (blue) of all 100 users of the PFL setting. In order from top to bottom: Random prior-$f_1$, Random prior-$f_2$, Learnt prior-$f_1$, Learnt prior-$f_2$.}
    \label{fig:histo}
\end{figure}

     \begin{table}[bt]
          \centering
          \caption{Results for the PFL scenario. $\ell^{0-1}$ corresponds to the 0-1 loss. The test error column is made on the test set of each user (10\% of local dataset).The
          Gen. Bound column gathers generalisation bounds. Each user bound is computed locally with $m_i=300$ for learnt prior, while $m_i=540$ for random prior.}
          \vspace{2mm}
          \begin{tabular}{|cc|ccc|}
          \hline
               \multicolumn{2}{|c|}{Setup}                                     & \multicolumn{3}{c|}{Gen. Bound $\ell^{0-1}$}                       \\ \hline 
               \multicolumn{1}{|c|}{Prior}  & Obj.                             & \multicolumn{1}{c|}{min}   & \multicolumn{1}{c|}{mean } & max   \\ \hline
               \multicolumn{1}{|c|}{Random} & \multicolumn{1}{|c|}{$f_{1}$}    & \multicolumn{1}{c|}{0,063} & \multicolumn{1}{c|}{0,680} & 0,847 \\ 
               \multicolumn{1}{|c|}{(us)}   & \multicolumn{1}{|c|}{$f_{2}$   } & \multicolumn{1}{c|}{0,075} & \multicolumn{1}{c|}{0,713} & 0,893 \\ \hline
               \multicolumn{1}{|c|}{Learnt} & \multicolumn{1}{|c|}{$f_{1}$}    & \multicolumn{1}{c|}{0,054} & \multicolumn{1}{c|}{0,112} & 0,222 \\ 
               \multicolumn{1}{|c|}{(us)}   & \multicolumn{1}{|c|}{$f_{2}$   } & \multicolumn{1}{c|}{0,052} & \multicolumn{1}{c|}{0,111} & 0,220 \\ \hline 
               \multicolumn{2}{|c|}{}       & \multicolumn{3}{c|}{Test Error $\ell^{0-1}$} \\ \hline 
               \multicolumn{1}{|c|}{Random} & \multicolumn{1}{|c|}{$f_{1}$}    & \multicolumn{1}{c|}{0}     & \multicolumn{1}{c|}{0,552} & 0,767 \\  
               \multicolumn{1}{|c|}{(us)}   & \multicolumn{1}{|c|}{$f_{2}$}    & \multicolumn{1}{c|}{0}     & \multicolumn{1}{c|}{0,588} & 0,833 \\ \hline
               \multicolumn{1}{|c|}{Learnt} & \multicolumn{1}{|c|}{$f_{1}$}    & \multicolumn{1}{c|}{0}     & \multicolumn{1}{c|}{0,050} & 0,183 \\ 
               \multicolumn{1}{|c|}{(us)}   & \multicolumn{1}{|c|}{$f_{2}$}    & \multicolumn{1}{c|}{0}     & \multicolumn{1}{c|}{0,044} & 0,150 \\ \hline 
          \end{tabular}
          \label{tab:PerezOrtiz2020personalised}
          \end{table}
     \Cref{tab:PerezOrtiz2020personalised} provides an overview of our results in the non-\iid case. It gathers, for both generalisation bounds and test error, the minimum, mean and maximum performance of all 100 users. The averaged performances are deteriorated compared to \Cref{tab:PerezOrtiz2020Ordered} for all settings as we consider a harder problem. It is worth noticing that our bounds are nonvacuous and that our algorithms with learnt priors benefit from bounds and test errors lower than 50\%. Indeed, if we do not learn the prior, we see from the distribution of errors in \Cref{fig:histo} that most users have a deteriorated bound and test errors. For learnt priors, the error distribution shows that all users enjoy a meaningful bound as well as sound performance. An interesting point is that the common prior distribution does not support all users uniformly as we can see in \Cref{fig:histo}. Indeed, our algorithms with random priors suffer from deteriorated bounds and test errors on average and the worst case, but approximately 15\% enjoy good test errors and 9\% benefit from theoretical guarantees lower than 40\%. Also, our algorithms with learnt priors enjoy test errors and generalisation guarantees lower than 20\% for all users. Furthermore, approximately half of the users benefit from test errors lower than $5\%$. This highlights the importance of the prior in this non-\iid setting. As the test set of each user is small (60 images), some users achieve a 0\% test error.

\subsection{Classification on CIFAR-10}
\label{sec:cifar10}
To evaluate the effectiveness of our algorithm in more challenging scenarios, we conduct experiments on the CIFAR-10 dataset. We followed a setup akin to \Cref{sec: fram-MNIST}, with minor adjustments. Specifically, we exploit a learned prior trained with 70\% of the dataset, the remaining being used for posterior estimation. Given the heightened complexity of the CIFAR-10 dataset, we involve Convolutional Neural Networks (CNNs) with 4 and 9 layers, denoted as CNNet4l and CNNet9l, respectively (these architectures also appear in \cite{perezortiz2021tight}, exploited as a baseline). To ensure enhanced performances, we also perform fine-tuning of hyperparameters related to FL such as local batch size and the number of local epochs, while keeping those associated with PAC-Bayes fixed to ensure relevant comparisons with the baseline.

We conducted a comprehensive grid search across a spectrum of hyperparameters to approximate the optimal model configurations. Specifically, we explored various combinations of local batch sizes $(1, 5, 10, 50)$ and local epoch counts $(1, 5, 10)$. Recognising that 100 rounds of training were inadequate for convergence, we extended the training duration to 300 rounds. Additionally, we adjusted the learning rate by a factor of 10 at round 200 to ease optimisation. From the resulting best configurations, we selected the most promising priors for further investigation. Subsequently, we fine-tuned the learning rate by exploring values between $5\times10^{-4}$ and $1\times10^{-3}$. Remarkably, we discovered that the optimal learning rate, as reported in \cite{perezortiz2021tight}, remains consistent between the centralised and federated settings. Furthermore, we conducted experiments to tune the dropout rate, revealing that a dropout rate of 0.2 yielded the best performance on the CIFAR-10 dataset, consistent with findings in \cite{perezortiz2021tight}. This reaffirms the applicability of centralised PAC-Bayes hyperparameters to the decentralised PAC-Bayes paradigm. Detailed results of these experiments, showing accuracies for each experiments on the batch size, epoch counts, learning rate and dropout rate are provided in Appendix \ref{sec:cifar10_expes}.

Our analysis revealed that the most effective hyperparameter settings were a local batch size of 5 for CNNet9l and 10 for CNNet4l, with a corresponding local epoch count of 1 for both architectures. Notably, increasing the number of epochs led to quicker convergence but marginally reduced accuracy. A learning rate of $5\times10^{-3}$ was optimal for both architectures, while a dropout rate of 0.2 was found to be the most effective.

We selected the priors with the highest accuracy from our grid search results (refer to the appendix for accuracy details). Subsequently, the posteriors were trained using identical hyperparameters as their corresponding priors. Additionally, we experimented with the KL penalty technique. Our findings are summarized in Table \ref{tab:CIFAR10} for comparison with the baseline results presented in the first row. Notably, both the prior and posterior performances are slightly inferior to the baseline. This discrepancy can be attributed to the inherent difficulty of federated learning compared to the batch setting. However, it is crucial to note that the generalisation bounds remain non-vacuous.

Among the configurations tested, the most promising outcome was observed with CNNet9l using a KL penalty of 1.0 in conjunction with $f_2$. This configuration achieved a generalisation bound and a test error of 34.5\% and 30.5\%, respectively, which is 10\% higher than the baseline for both metrics. This a similar conclusion than in \Cref{sec: expe-MNIST}, highlighting the price to pay to switch from batch to FL. Surprisingly, the KL penalty trick did not yield an improvement in the generalisation bound, contrary to \Cref{sec: expe-MNIST}. This is possibly linked to the intrinsic complexity of CIFAR-10. In particular, the inadequacy of the prior may necessitate further optimisation during posterior training, potentially causing the posterior to diverge significantly from the prior distribution. Consequently, the KL term increases post-training, warranting a more substantial penalty.

\begin{table}[]
     \begin{center}
     \begin{tabular}{|c|c|c|c|c|c|c|c|c|}
     \hline
     Model   & Obj. & $\beta$ & $\epsilon$ & rounds & Bound & Test Error & KL Div & Prior Test Error  \\ \hline
     \multicolumn{9}{|c|}{\cite{perezortiz2021tight}}                                                      \\ \hline
     CNNet9l & $f_1$ & 250       & 100        & "1" & 0.237 & 0.216      & <0.0001  & 0.217              \\ 
     (baseline)& $f_2$ & 250       & 100        & "1" & 0.250 & 0.214      & 0.003    & 0.217              \\ \hline
     \multicolumn{9}{|c|}{100 users - GenFL - KL Penalty=0.1}                                              \\ \hline
     CNNet4l & $f_1$ & 10        & 1          & 300 & 0.471 & 0.388      & 0.012  & 0.329               \\ 
     (us)    & $f_2$ & 10        & 1          & 300 & 0.469 & 0.391      & 0.011  & 0.329               \\ \hline
     CNNet9l & $f_1$ & 5         & 1          & 300 & 0.393 & 0.303      & 0.019  & 0.274               \\ 
     (us)    & $f_2$ & 5         & 1          & 300 & 0.396 & 0.304      & 0.018  & 0.274               \\ \hline
     \multicolumn{9}{|c|}{100 users - GenFL - KL Penalty=1.0}                                               \\ \hline
     CNNet4l & $f_1$ & 10        & 1          & 300 & 0.466 & 0.390      & 0.011  & 0.329               \\ 
     (us)    & $f_2$ & 10        & 1          & 300 & 0.458 & 0.391      & 0.009  & 0.329               \\ \hline
     CNNet9l & $f_1$ & 5         & 1          & 300 & 0.386 & 0.302      & 0.014  & 0.274               \\ 
     (us)    & $f_2$ & 5         & 1          & 300 & 0.345 & 0.305      & 0.003  & 0.274               \\ \hline
     \end{tabular}
     \caption{Table displaying the results for the CIFAR-10 dataset alongside the baseline (batch setting) presented in the first row. 'Prior Test Error' represents the 0-1 error of the prior on the test set. The symbol $\beta$ denotes the batch size of clients, while $\epsilon$ indicates the number of local epochs on each round.}
     \label{tab:CIFAR10}
     \end{center}
\end{table}

\section{Discussion}
\label{sec:discussion}
In this work, we propose a novel algorithm for FL in two different settings: FL-SOB, which allows to exploit a global generalisation guarantee while keeping data separated; as well as PFL, which only involves an \iid assumption for each user's dataset. Our work raises two questions: (a) is it possible to remove the \iid assumption?(b) is it possible to maintain a global generalisation guarantee, even in the personalised setting?   To answer (a), a line of work first initiated by \cite{kuzborskij2019efron} (for \iid data) and continued by \cite{haddouche2023pacbayes,ChuggWangRamdas2023,jang2023tighter}  (for non-\iid ones) focuses on PAC-Bayes bounds valid for data distribution with bounded variances. In the PFL setting, this could provide novel generalisation bounds without assuming each user possesses an \iid dataset. About (b), the recent work of \cite{sefidgaran2023fed} provides elements of answer: they derive a general PAC-Bayesian bound holding for the classical FL setting \emph{for all users simultaneously} involving explicitly the number of users and rounds. This allows fruitful theoretical interpretations (especially on the number of rounds involved during the FL training), but leads to vacuous generalisation guarantees for classification task with a federated SVM. Following another route, based on PAC-Bayes methods for meta-learning, \cite{boroujeni2024personalized} provides a novel FL algorithm for PFL derived from an original theoretical result with strong performances. However, their approach involves distributions on distributions spaces, giving their method a potentially high time complexity, they are also unable to compute nonvacuous generalisation guarantees. This constrasts with our results, even for the personalised setting, at the cost of considering generic PAC-Bayesian bounds, not explicitly tailored for federated learning. Establishing a PAC-Bayes bound designed for FL and leading to a non-vacuous generalisation guarantees remains an open challenge that we aim to address in a future work.





\paragraph{Acknowledgements}
M. Haddouche is supported by the European Union (ERC grant DYNASTY
101039676). The French government partly funded this work under the management of Agence
Nationale de la Recherche as part of the “France 2030” program, reference ANR-23-IACL-0008
(PR[AI]RIE-PSAI). B. Guedj acknowledges partial support from the French National Agency for
Research, through the programme “France 2030” and PEPR IA on grant SHARP ANR-23-PEIA0008.

\bibliography{main}
\bibliographystyle{abbrvnat}

\appendix


\subsection*{Appendix 1 Approximating the inverted $kl$}

\label{sec: kl_calculus}
To approximate the inverted kl divergence, we re-use the technique presented in \cite[Appendix A]{DziugaiteRoy2017}. As there is no closed form formula for $\mathrm{kl}^{-1}(q \mid c)$, we approximate it via root-finding techniques. For all $q \in$ $(0,1)$ and $c \geq 0$, define $h_{q, c}(p)=\operatorname{KL}(q \| p)-c$. Then $h_{q, c}^{\prime}(p)=\frac{1-q}{1-p}-\frac{q}{p}$. Assuming we possess a good anough initial estimate $p_0$ of a root of $h_{q, c}(\cdot)$, we can obtain improved estimates of a root via Newton's method:
$$
p_{n+1}=\mathrm{N}\left(p_n ; q, c\right) \text { where } \mathrm{N}(p ; q, c)=p-\frac{h_{q, c}(c)}{h_{q, c}^{\prime}(p)} \text {. }
$$
This suggests the following approximation to $\mathrm{kL}^{-1}(q \mid c)$ :
1. Let $\tilde{b}=q+\sqrt{\frac{c}{2}}$.
2. If $\tilde{b} \geq 1$, then return 1 .
3. Otherwise, return $\mathrm{N}^k(\tilde{b})$, for some integer $k>0$.

\subsection*{ Appendix 2 Additional details for \Cref{sec:personalised_fl}}
\label{sec: details_personalised}
We provide here more details about the procedures of \cref{sec:personalised_fl}. 

\vspace{1mm}
\textbf{PAC-Bayes learning objectives} As stated in the main documents, our learning objectives here are mainly similar to those in \Cref{eq:obj_classic_ordered,eq:obj_quad_ordered}. At the variation that now, the KL divergence is regularised by $m_i$ for the user $i$ instead of the common $m$. This comes from the fact that now, each user does not try to optimise proxys of a common global generalisation goal but only its personal McAllester bound, depending only on its data.

\begin{align*}
f_{1}(\S_i) & =  \hat{\Risk}_{\S_i}(Q) + \sqrt{ \frac{\KL(Q\|P) + \ln{\frac{2\sqrt{m_i}}{\delta}}}{2m_i}} \\
f_{2}(\S_i) & = \left(\sqrt{\hat{\Risk}_{\S_i}(Q)+\frac{\mathrm{KL}\left(Q \| P\right)+\log \left(\frac{2 \sqrt{m}}{\delta}\right)}{2 m_i}} +\sqrt{\frac{\KL\left(Q \| P\right)+\log \left(\frac{2 \sqrt{m}}{\delta}\right)}{2 m_i}}\right)^2
\end{align*}

More precisely we state explicitly the personalised algorithm we use in \Cref{alg:p-FL_alg}. This algorithm shows that each user sacrifices half of its data in phase 1 to learn jointly a prior through \textsc{GenFL}. In phase 2, we re-use the ClientUpdate procedure of \cref{alg:GenFL} to personalise the prior to each client through the PAC-Bayesian learning goal $f$ (being $f_{1}$ or $f_{2}$ in practice).

\begin{algorithm}[t]
\caption{ PFL algorithm with PAC-Bayesian personalisation step. The K clients are indexed by k; B is the local minibatch size, E is the number of local epochs, $\eta$ is the learning rate $f$ PAC-Bayes objective, $\mu_{\text{prior}}$ prior center parameters, $\sigma_{\text{prior}}$ prior scale hyper-parameter, $\delta$ confidence parameter}\label{alg:p-FL_alg}
\begin{algorithmic}[1]
\State $w_0 \leftarrow (\mu_{\text{prior}}, \sigma_{\text{prior}})$
\State $\forall k,\S_k = \S_k^{1}\cup \S_k^{2}$ \Comment{splitting datasets in half}
\State \textbf{Step 1: learning the prior.}
\State $\S_\P \leftarrow \cup_{k=1}^{K} \frac{1}{2} \S_k^{1}$ \Comment{each user involves half of its dataset.}
\State $\mu_\P \leftarrow \textsc{GenFL}(\mu_{prior},\S_\P)$
\State $w_1 \leftarrow (\mu_\P, \sigma_{prior})$ \Comment{Prior is constructed.}
\State \textbf{Step 2: Personalisation step.}
\For{each client $k= 1, \cdots K$}
\State $w_{\Q}^k \leftarrow ClientUpdate(k, w_1, \frac{m_k}{2})$ 
\EndFor
\State Return $(w^{k}_{\Q})_{k=1\cdots K}$
\State
\State \textbf{ClientUpdate(k, w, m):}
\State $\mathcal{B} \leftarrow (\text{split } \S_k^{2} \text{ into batches of size B})$
\For{each local epoch $e= 1, 2, \cdots, E$}
\For{each local minibatch $b \in B$}
\State $w^k_s \leftarrow \mu^k + \sigma^k \odot \mathcal{N}(0,1)$ \Comment{Reparam. trick}
\State $w^k \leftarrow w^k - \eta \nabla_w f_{m, \delta, \mu_{\text{prior}}, \sigma_{\text{prior}}}(w^k_s; b)$
\EndFor
\EndFor
\State \textbf{return} $w^k$
\Ensure Personalised distributions $w_\Q^k \Leftrightarrow \Ncal(\mu_\Q^k,\sigma_\Q^k)$
\end{algorithmic}
\end{algorithm}

For the sake of completeness, we also precise in \Cref{alg:p-FLBound} how we calculate the personalised generalisation bounds once the PFL training of \Cref{alg:p-FL_alg} is performed. 

\begin{algorithm}[t]
\caption{\textsc{FedBound}. The K clients are indexed by k; $f$ PAC-Bayes objective, prior distribution $\mu_{\P}$, $\sigma_{\text{prior}}$; posterior $\Ncal(\mu_{T}$ output of \Cref{alg:p-FL_alg}, $\delta$, $\delta^{'}$ confidence parameters, $n$ number of MC sampling}
\label{alg:p-FLBound}
\begin{algorithmic}[1]
\State \textbf{Server executes:}
\State $\P = \mathcal{N}(\mu_{\P}, \sigma_{\text{prior}})$ \Comment{ Learned prior from \cref{alg:p-FL_alg}}
\State $\Q = \mathcal{N}(\mu_{T}, \sigma_{T})$ \Comment{Posterior (learned)}
\For{each client $k \in K$ \textbf{in parallel}}
\State $error^{k} \leftarrow ClientMCSampling(k, w_t, m_k/2)$
\State $KL\_inv \leftarrow \emph{kl}^{-1} \left(error^k \mid \frac{1}{n} \ln(\frac{2}{\delta^{'}})\right)$
\State $\texttt{Up-bound}_k \leftarrow \emph{kl}^{-1} \left(KL\_inv \mid \frac{\KL(Q\|P) + \ln{\frac{2\sqrt{m}}{\delta}}}{m}\right)$
\EndFor

\State
\State \textbf{ClientMCSampling(k, w, m):}
\For{each MC sampling $i= 1, 2, \cdots, n$}
\State $W^k_i \sim \Q$ \Comment{Sample weights from the posterior}
\State $error^{k}_i \leftarrow \hat{\Risk}_{\S_k^2}(W^k_i)$ \Comment{local empirical risk}
\EndFor
\State $error^{k} \leftarrow = \frac{1}{n} \sum_{i=1}^{n} error^{k}_i$
\State \textbf{return} $error^{k}$
\Ensure $\LP\texttt{Up-bound}_k\RP_{k=1\cdots K}$ each valid with probability $1- \delta - \delta^{'}$
\end{algorithmic}
\end{algorithm}

\Cref{alg:p-FLBound} simply computes each PAC-Bayesian associated fort each client. Note that here $\S_k^2$ denotes the halve of $\S_k$ which has not been used to train the prior distribution.

\subsection*{Appendix 3 Additional experiments }
\label{sec:additional_expes}

\subsubsection*{FL-SOB scenario variability}
\label{sec: additional_pfl}

\textbf{Variability of the FL-SOB scenario.}
We provide additional experiments to show the variability of the learning procedure. We run 10 times our experiments with different seeds using \Cref{sec:expes}'s setup to assess the robustness of our approach. We report the results in \Cref{tab:PerezOrtiz2020OrderedStd}.

\begin{table}[bt]
     \caption{ Results for the FL-SOB scenario with 10 differents seeds.
     We have $\ell^{0-1}$ corresponds to the 0-1 loss. The test error column is made on the test set of MNIST. The
     Bound column corresponds to the generalisation bounds, computed with \cref{alg:FedBound}. The $KL/m$ column corresponds to the KL divergence term in the bound divided by $m=60000$ in data-free prior or $m=30000$ data-dependent prior.
     In addition, we computed the mean and standard variation (std.) for 10 differents seeds for all the metrics.}
     \label{tab:PerezOrtiz2020OrderedStd}
     \vspace{2mm}
     \centering
     \begin{tabular}{|cc|cc|cc|cc|}
     \hline
          \multicolumn{2}{|c|}{Setup}              & \multicolumn{2}{c|}{$\ell^{0-1}$ Bound}  & \multicolumn{2}{c|}{$\ell^{0-1}$ Test Err.}  & \multicolumn{2}{c|}{$KL/m$}      \\ \hline 
          \multicolumn{1}{|c|}{Prior}  & Obj.      & \multicolumn{1}{c|}{Mean}  & Std.        & \multicolumn{1}{c|}{Mean} & Std.             & \multicolumn{1}{c|}{Mean} & Std. \\ \hline
          \multicolumn{8}{|c|}{100 users - GenFL - No KL Penalty} \\ \hline
          \multicolumn{1}{|c|}{Random} & $f_{1}$   & \multicolumn{1}{c|}{0.4188} & 0.0041     & \multicolumn{1}{c|}{0.2609} & 0.0036         & \multicolumn{1}{c|}{0.0387}    & 0.0006 \\ 
          \multicolumn{1}{|c|}{(us)}   & $f_{2}$   & \multicolumn{1}{c|}{0.4105} & 0.0037     & \multicolumn{1}{c|}{0.2531} & 0.0041         & \multicolumn{1}{c|}{0.0404}    & 0.0007 \\ \hline
          \multicolumn{1}{|c|}{Learnt} & $f_{1}$   & \multicolumn{1}{c|}{0.0397} & 0.0004     & \multicolumn{1}{c|}{0.0309} & 0.0010         & \multicolumn{1}{c|}{$<$0.0001} & $<$0.0001 \\ 
          \multicolumn{1}{|c|}{(us)}   & $f_{2}$   & \multicolumn{1}{c|}{0.0400} & 0.0003     & \multicolumn{1}{c|}{0.0301} & 0.0008         & \multicolumn{1}{c|}{0.0003}    & $<$0.0001 \\ \hline
    \end{tabular}
\end{table}

\subsubsection*{Additional experiments to compare with \cite{DziugaiteRoy2017}}
\label{sec: comp_dziugaite}
To prove the flexibility of \textsc{GenFL} with respect to various state-of-the-art numerical experiments, we modify \textsc{GenFL} to make it in line with \cite{DziugaiteRoy2017}. They exploited a different prior than \Cref{sec:expes} which is explained in the next section.
\\
\subsubsection*{Setup}

\textbf{Dataset Partition}
To build a \iid FL setup, we did as in \Cref{sec:expes}, we
consider the case where each user has exactly the same
number of samples per class. We partition MNIST as
follows: we fix the number of users to 100. Then, each
user receives a dataset size of 540, each class having
54 images.

\vspace{1mm}
\textbf{Random initialization of the prior}
The prior is $P=\mathcal{N}(w_0, \lambda Id)$, and $\lambda$ is a  parameter discretised on a grid $\lambda = (c\exp(-j/b))_{j\in\N}$ with $c=0.1,b=100$.
We sampled $w_0 \sim \mathcal{N}(0, \sigma)$ and then truncated to $[-2\sigma,2\sigma]$, we used $\sigma=0.04$.

\vspace{1mm}
\textbf{SGD on centered parameters of the posterior}
\cite{DziugaiteRoy2017} proposed to learn the centered parameters of the posterior $w$ via SGD on training set. This step allows better results than learning the posterior directly.

\vspace{1mm}
\textbf{Learn posterior from the SGD}
The posterior is initalised to $Q=\mathcal{N}(w, Diag(|w|))$ ($w$ coming from the previous SGD). This posterior and $\lambda$ are optimised on the learning objective $f_1$.

\vspace{1mm}
\textbf{Bounds parameters}
Bounds were computed with confidence parameters $\delta = 0.05$, $\delta^{'} = 0.01$ and with $n=150000$ monte carlo samples.

\subsubsection*{Results}

\begin{table}[htbp]
    \centering
    \caption{Results for the FL-SOB scenario on Dziugaite setting. $\ell^{0-1}$ corresponds to the $0-1$ loss. The test error column is made
    on the test set of MNIST. The Bound column corresponds to the generalisation bounds, computed with \Cref{alg:FedBound} The $KL/m$ column corresponds to the KL divergence term in the bound divided by m.}
    \begin{tabular}{|c|c|c|c|}
    \hline
    Metrics         & GenFL  & GenFL  & Dziugaite   \\ \hline
    Clients number  & 100     & 1       & "1"       \\ \hline
    Trainset Size   & 54000   & 54000   & 60000     \\ \hline
    Train Error     & 0.013   & $\sim$0 & $\sim$0   \\ \hline
    Test Error      & 0.019   & 0.018   & 0.016     \\ \hline
    SNN Train Error & 0.051   & 0.039   & 0.028     \\ \hline
    SNN Test Error  & 0.051   & 0.043   & 0.033     \\ \hline
    PAC-Bayes Bound & 0.241   & 0.175   & 0.186     \\ \hline
    KL divergence   & 7039    & 4629    & 6534      \\ \hline
    \end{tabular}
\label{tab:DziugaiteRoy2017}
\end{table}

\textbf{FL-SOB on Dziugaite Setting.}
\Cref{tab:DziugaiteRoy2017} gathers our results for \textsc{GenFL} applied with $f_1$ alongside \textsc{FedBound} with prior described above. We provide our metrics with 100 clients, 1 client and compare ourselves with \cite{DziugaiteRoy2017} (1 client). Our FL algorithm benefits from nonvacuous theoretical guarantees and test errors. The SNN train and test errors of the posterior almost match \cite{DziugaiteRoy2017} by 2\%. The PAC-Bayes bound is slightly deteriorated (0.241 vs 0.186), we interpret this as the cost to move from batch to federated learning.

\textbf{Variability of the FL-SOB scenario.}
The results in \Cref{tab:PerezOrtiz2020OrderedStd} are very similar to the one in \Cref{tab:PerezOrtiz2020Ordered}.
Standard deviations are 100 times smaller than the means. It shows that the learning procedure is stable.

\subsection*{Appendix 4 Additional experiments on CIFAR-10}
\label{sec:cifar10_expes}

\begin{figure}[htbp!]
    \centering
    \includegraphics[width=0.8\textwidth]{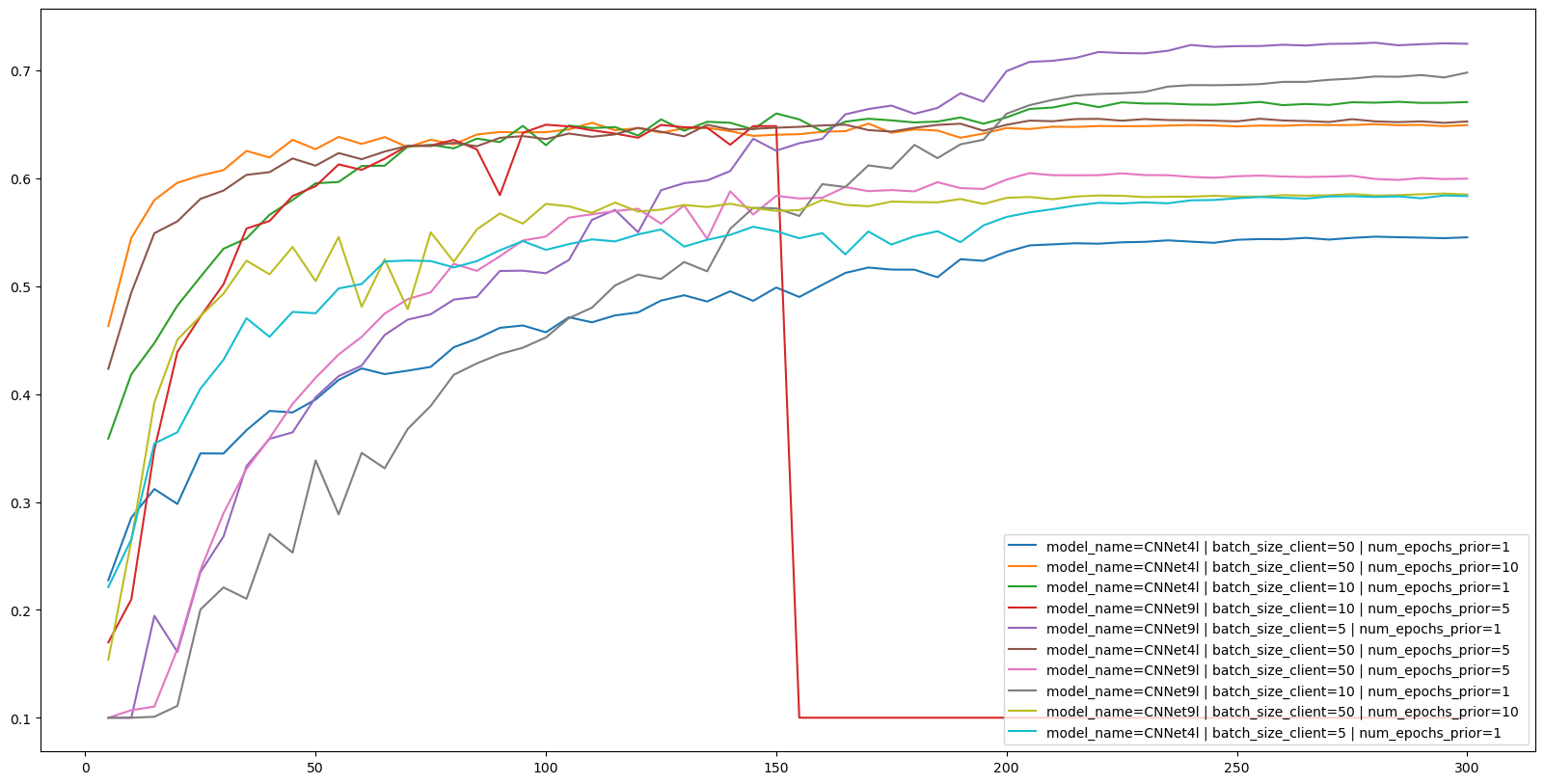}
    \caption{Grid search results showcasing the accuracies of various priors on the CIFAR-10 test set. Prior training involved two CNN architectures with differing local batch sizes (1, 5, 10, 50) and local epoch counts (1, 5, 10) across 300 rounds, with a learning rate reduction by a factor of 10 at round 200. Only models achieving accuracies exceeding 50\% are depicted.}
    \label{fig:cifar10_results_nb_updates}
\end{figure}

In the CIFAR-10 experiments discussed in \Cref{sec:cifar10}, we adopt a comparable setup with identical hyperparameters. Through hyperparameter optimization, we determine the optimal number of local updates, learning rate, and dropout rate for training the prior. The outcomes are presented in \Cref{fig:cifar10_results_nb_updates}, \Cref{tab:cifar10_results_lr}, and \Cref{tab:cifar10_results_dropout}, respectively.

\textbf{Varying the number of updates.}

\textbf{Varying the learning rate.}

For each CNN, we selected the optimal configuration regarding local batch size and local number of epochs, then explored various learning rates. The outcomes are summarized in \Cref{tab:cifar10_results_lr}. Notably, the learning rate demonstrates a substantial impact on model test error, with the most favorable values observed at 0.003 and 0.005 for both models.

\begin{table}
    \centering
    \begin{tabular}{|c|c|c|c|c|}
    \hline
    model name & batch size & num epochs & learning rate & Test Error \\ \hline
    CNNet9l	   & 5	  & 1	& 0.0030 & 0.271 \\ \hline
    CNNet9l	   & 5	  & 1	& 0.0050 & 0.288 \\ \hline
    CNNet9l	   & 5	  & 1	& 0.0010 & 0.316 \\ \hline
    CNNet4l	   & 10  & 	1	& 0.0030 & 0.327 \\ \hline
    CNNet4l	   & 10  & 	1	& 0.0050 & 0.329 \\ \hline
    CNNet4l	   & 10  & 	1	& 0.0010 & 0.334 \\ \hline
    CNNet4l	   & 10  & 	1	& 0.0007 & 0.350 \\ \hline
    CNNet4l	   & 10  & 	1	& 0.0005 & 0.369 \\ \hline
    CNNet9l	   & 5	  & 1	& 0.0007 & 0.374 \\ \hline
    CNNet9l	   & 5	  & 1	& 0.0005 & 0.452 \\ \hline
    \end{tabular}
    \caption{Results for the learning rate experiment on CIFAR-10 dataset. We report the 0-1 error on the test set of the models after 300 rounds of training. The learning rate values are 5e-4, 7e-4, 1e-3, 3e-3, 5e-3. The learning rate is decreased by a factor of 10 at round 200.}
    \label{tab:cifar10_results_lr}
\end{table}

\textbf{Varying the dropout rate.}
We further experimented with varying the dropout rate for each CNN and present the outcomes in \Cref{tab:cifar10_results_dropout}. Notably, the dropout rate exhibits a considerable influence on model test error, with optimal values observed at 0.2 for both models.

\begin{table}
    \centering
    \begin{tabular}{|c|c|c|c|c|c|c|c|c}
    \hline
    model name	& batch size &	num epochs & dropout rate & Test Error \\ \hline
    CNNet9l	& 5 &	1 &	0.2 & 0.267 \\ \hline
    CNNet4l	& 10 &	1 &	0.2 & 0.328 \\ \hline
    CNNet9l	& 5 &	1 &	0.3 & 0.331 \\ \hline
    CNNet4l	& 10 &	1 &	0.4 & 0.339 \\ \hline
    CNNet4l	& 10 &	1 &	0.3 & 0.341 \\ \hline
    CNNet4l	& 10 &	1 &	0.5 & 0.360 \\ \hline
    CNNet9l	& 5 &	1 &	0.4 & 0.434 \\ \hline
    CNNet9l	& 5 &	1 &	0.5 & 0.574 \\ \hline
    \end{tabular}
    \caption{Results for the dropout rate experiment on CIFAR-10 dataset. We report the best 0-1 error on the test set of the models after 300 rounds of training. The dropout rate values are 0.2, 0.3, 0.4, 0.5.}
    \label{tab:cifar10_results_dropout}
\end{table}

\end{document}